# Multi Task Deep Morphological Analyzer: Context Aware Neural Joint Morphological Tagging and Lemma Prediction

SAURAV JHA and AKHILESH SUDHAKAR, Independent Researchers
ANIL KUMAR SINGH, Indian Institute of Technology (BHU)

The ambiguities introduced by the recombination of morphemes constructing several possible inflections for a word makes the prediction of syntactic traits in Morphologically Rich Languages (MRLs) a notoriously complicated task. We propose the Multi Task Deep Morphological analyzer (MT-DMA), a character-level neural morphological analyzer based on multitask learning of word-level tag markers for Hindi and Urdu.[1] MT-DMA predicts a set of six morphological tags for words of Indo-Aryan languages[2]: Parts-of-speech (POS), Gender (G), Number (N), Person (P), Case (C), Tense-Aspect-Modality (TAM) marker as well as the Lemma (L) by jointly learning all these in one trainable framework. We show the effectiveness of training of such deep neural networks by the simultaneous optimization of multiple loss functions and sharing of initial parameters for context-aware morphological analysis. Exploiting character-level features in phonological space optimized for each tag using multi-objective genetic algorithm, our model establishes a new state-of-the-art accuracy score upon all seven of the tasks for both the languages. MT-DMA is publicly accessible: code, models and data are available at MT-DMA's Github repository[3].

## 1 INTRODUCTION

Morphological analysis [52, 38] refers to the task of assigning a set of well-defined morphological tags (hereby, referred to as tags) and a lemma (root) to the tokens of a language by studying a range of syntactic traits such as inflection, derivation, combining forms, and cliticization [45]. As a forerunner to a range of complex tasks (word-sense disambiguation, spell checker, machine translation, dependency parsing, etc.), morphological analysis remains an important research idea in the field of NLP. For instance, in machine translation of Low-Resource Languages (LRLs)[4] that are also MRLs, morphological analysis helps in sparsity reduction by breaking down each word into its tags and lemma. The translation framework then needs to translate only the lemma (the set of all lemmas is much smaller than the set of all word forms for most MRLs), and then use the tags to re-generate the exact inflection of the word in the target language.

In this work, we consider six universal tags that characterize the morphology of every word in Indic languages [73]: Parts-of-speech (POS), Gender (G), Number (N), Person (P), Case (C), Tense-Aspect-Modality (TAM)[5] as well as the Lemma (L). Prior to the advent of deep learning,

---

[1]Both Hindi and Urdu are Indo-Aryan languages that are MRLs and are often clubbed together as Hindustani languages.
[2]The family of languages spoken most widely in the Indian subcontinent including India, Nepal, Pakistan, Bangladesh, and Srilanka.
[3]https://github.com/Saurav0074/morph_analyzer
[4]Our definition of LRLs include languages lacking adequate online corpora in lieu of those having low density.
[5]The tag covering the expression of tense (T), aspect (A), and modality (M).

---

Authors' addresses: Saurav Jha, mail@sauravjha.com.np; Akhilesh Sudhakar, akhileshs.s4@gmail.com, Independent Researchers; Anil Kumar Singh, Indian Institute of Technology (BHU), nlprnd@gmail.com.

two widely accepted methods for such characterization were: (a) dictionary-based approach using look-up tables to retrieve grammatical information of matching words in a language [65, 75], and (b) rule-based approach leveraging a set of inflection rules to first retrieve the base form of a word followed by its tags based on suffix and prefix matches [34, 42]. Having evolved as successors to the sequential phonological rewrite rules and the parallelized two-level morphology rules [47], these approaches, however, continue to suffer from the problem of laborious manual design of rules and dictionaries. In addition, they (a) demand high memory requirements while (b) remaining highly language specific. Recently, statistical learning methods have proved to be more promising for the task [51, 70].

## 1.1 Deep Neural Architectures

The universality of Artificial Neural Networks (ANN) in computing functions[6] has opened up possibilities of adapting these to the task of learning morphology of languages [54, 49, 80]. Deep learning models represent each token as a dense low-dimensional real valued vector (a.k.a. embedding) wherein each dimension captures a latent feature of the token. In recent times, the CNN-RNN model [84, 28], which leverages the strength of both Convolutional Neural Networks (CNNs) and Recurrent Neural Networks (RNNs) for sequence prediction problems with spatial inputs, have shown promising results when used in morphological understanding. The translation and rotation invariance of CNNs over inputs with local consistency makes them a good choice for the extraction of local features such as n-grams from different positions of text through the use of filters. Combining this with the ability of RNNs to capture long-term dependencies, such models stand to be effective candidates for modeling text representation [77, 84, 76, 66, 17].

## 1.2 Morphology of Hindustani Languages

Originated as a means of early communication between the speakers of Khariboli (the Delhi regional dialect of medieval Indo-Aryan Apabhramsa), and the 13[th] century Mughal officials and other Middle Eastern traders, *Hindustani* has since been the *lingua franca* of Northern India and Pakistan, with Hindi and Urdu being its modern-day descendants. Like other Indo-Aryan languages, both Hindi and Urdu possess morphological richness, thus causing modification of words via inflection to significantly outnumber those via agglutination. A word form in Hindi may have over 40 morphological analyses [31]. Table 1 shows one such example of multiple valid context-independent analyses for the hindi word पूरे. On considering the context in which a particular word is used in a sentence, only one such analysis holds correct. Though the inflectional morphology of Hindi is not agglutinative, the derivational suffixes are. This further leads to an explosion in the number of inflectional lemma forms [69]. As also mentioned in Sudhakar and Singh [72], one of the reasons for our focus on Hindi is that it has a wide coverage of speaking population with over 260 million speakers across 5 countries[7], and is the fifth most spoken language in the world. Urdu too, is a widely spoken language, and is the 21[st] most spoken first language in the world with 66 million speakers [58].

*1.2.1 Complexity of Hindustani Morphology.* Inspite of the widespread usage of both the languages, far less work has been done on building morphological analyzers for either language (see Section 5) owing to a number of limitations. First, the morphological outreach of Hindi/Urdu beyond the lexical level [53] requires any such analyzer to exploit the agreement of tag markers across grammatical categories so as to achieve morphological distinctions. While these agreements arise out of analytic, synthetic, and agglutinative word formation techniques distinguishing

---

[6]http://neuralnetworksanddeeplearning.com/chap4.html
[7]https://www.ethnologue.com/statistics/size

| lemma | Category | Gender | Number | Person | Case | TAM |
|---|---|---|---|---|---|---|
| pUrA | adj | m | sg | - | o | - |
| pUrA | adj | m | pl | - | d | - |
| pUrA | adj | m | pl | - | o | - |
| pUrA | n | m | pl | 3 | d | 0 |
| pUrA | n | m | sg | 3 | o | 0 |
| pUra | v | any | sg | 2 | - | ए |
| pUra | v | any | sg | 3 | - | ए |
| pUra | v | m | pl | any | - | या |

Table 1. Morphological analyses of the word पूरे [pUre ] adapted from Sudhakar and Singh [72]

: '-' *indicates that the feature is not applicable and an 'any' indicates that it can take any value in the domain for that feature.*

between concatenative and non-concatenative formation rules [15], they strongly imply that a restricted set of tags make appropriate markers for different POS elements (i.e., G, N, and C for noun and adjective, C for pronoun, TAM for verb, and morphological invariability for the rest). Second, as mentioned by Davison [27], syntactic properties such as post-positional case marking, split ergative subject marking (required in case of transitive verbs, optional for others, and prohibited on the subjects of some bivalent verbs), complex predicates (formed from a variety of sources - Sanskrit, Persian, Arabic, and English) as well as the presence of several nearly synonymous doublets (such as simple verbs and complex predicates) inject a range of ambiguities.

### 1.3 Multi-Task Learning (MTL) of Morphological Structure

Unlike morphologically poor languages (e.g., English) wherein the restricted agreement among words require the analysis of relatively few features [30], MRLs show complex agreement patterns between POS, G, N, P, C, TAM, and L (see Appendix A). For instance, the dependency of TAMs on POS tag sequences have previously been exploited by Malladi and Mannem [51] through the use of language-specific heuristics. The deep-seated dependencies among these tags imply that the knowledge required to learn either could be useful in learning the others as well. Based upon this implication, we frame the overall task of learning the six tag groups as a multi-task learning objective - a well-established machine learning paradigm to learn a task using other related tasks [16].

Further, Chen et al. [18] show that the semantic composition functions of a language remain preserved across different tasks, at least on meta-knowledge level. We extend the same argument from the semantic space to the morphological space. MTL-based frameworks are known to perform inherent data augmentation by averaging the noise patterns between multiple tasks, extend the focus to the features relevant to a task while learning important features through an auxiliary task, and reduce overfitting by introducing an inductive bias.

This paper introduces a multi-task learning framework based upon two widely used architectures: (a) the aforementioned CNN-RNN model for predicting the POS, G, N, P, C, and TAM, and (b) an attention-based sequence-to-sequence model for predicting the L of Hindi/Urdu words. The multi-task deep morphological analyzer (MT-DMA) model uses the hard parameter-sharing strategy [16] to jointly learn to predict all seven tasks. We exploit the implicit relatedness among the tasks, by utilizing the knowledge acquired in learning the representations of either of the tags in predicting the others. In doing so, we present a rigorous exploration of fine-tuned linguistic features adapted to both the languages.

### 1.4 Our Contributions

The MT-DMA model differs from the existing methods in that: (a) Our work does not require prior linguistic knowledge of a specific language to take advantage of inter-tag agreements, since our framework learns these implicitly. (b) We leverage a wider context information of the input word up to a window of four to exploit the *subject-object-verb* (SOV) order agreement [13] among the words[8], thus offering substantial advantage over longer sentences. (c) While current multi-task morphology learning strategies utilize one [26], two [44] or three [81] related auxiliary tasks to improve upon a main task, we present a first use-case of the joint learning of seven correlated and equally weighted tasks, and show how the gains acquired by one can mutually benefit the others. (d) Our work offers new insight on the less-explored phono-morphological interface by using phonological features to enhance a traditionally lexical task while characterizing the tag-specific features via multi-objective optimization.

## 2 PROBLEM FORMULATION

We formulate morphological analysis as a multi-task learning problem. Formally, MT-DMA learns a shared representation of features across the six morphological tags in addition to learning separate feature representation for lemma of a word by minimizing the joint categorical cross-entropy loss functions for all seven tasks. Let, $i = 1,2,...,N$ be the individual observation instances, $c = 1,2,...,C$ the classes, and $j = 1,2,...,t$ the number of tasks in hand (in our case, $t=7$), then the combined cross-entropy loss function of the model can be stated as:

$$L_\theta = \sum_{j=0}^{6} \lambda_j (-\frac{1}{N} \sum_{i=1}^{N} \sum_{c=1}^{C} 1_{y_i \in C_c} \log p_{model}[y_i \in C_c]) \quad (1)$$

where, $p_{model}[y_i \in C_c]$ is a *C*-dimensional vector denoting the probability of $i^{th}$ observation belonging to the $c^{th}$ category.

The parameter vector $\theta$ in equation (1) remains shared among all seven softmax layers thus allowing the model to represent morphological regularities across the tasks through a common embedding space, while facilitating the transfer of knowledge among them. Further, the weight factors $\{\lambda_0, \lambda_1, ..., \lambda_6\}$[9] serve to check the discounts imparted upon the individual loss of the tasks while the model adjusts its parameters to minimize the overall loss $L_\theta$. For our task, we keep two distinct sets of such weights, i.e., $\{\lambda_L\}$ and $\lambda = \{\lambda_{POS}, \lambda_G, ..., \lambda_{TAM}\}$. As a heuristic for searching a good approximation to these, we ensure that the values of $\lambda$ and $\lambda_L$ sum together to 1. It is worth mentioning that we do not treat any of the tasks as auxiliary.

## 3 SYSTEM DESCRIPTION

This section presents the architecture and the working of MT-DMA. Section 3.1 gives an overview of the working of the model, and is backed by sections 3.2 and 3.3, explaining in detail about the components of the model. We also describe the strategies that we employ in training the shared parameters of the model in section 3.4.

### 3.1 Working in a nutshell

As depicted in Figure 1, our system integrates two components: (a) the **Tag Predictor**, and (b) the **Lemma Predictor**. The input to the model comprises the current word *'wrd'* and a sequence of up to (2*CW) words, where CW stands for the size of the Context Window. Best results are obtained with CW = 4. The final output consists of the predicted set of tags - POS, G, N, P, C, and TAM, yielded by the tag predictor, as well as the Lemma (L), generated by the lemma predictor for *wrd*. Given the recent success of character-level embeddings in morphology, their ability to handle Out-of-Vocabulary (OOV) words [79] and reduced vocabulary size, *i.e., determined by the total size of alphabets*[10], we employ character-level embeddings to capture intra-word morphological and shape information. The tag and the lemma predictors share a character embedding layer that represents the characters of the input words using a 64-dimensional vector.[11]

---
[8]Both Hindi and Urdu, besides other Indo-Aryan languages, show the SOV agreement. For instance, verbs in Hindi change forms according to N, G, P, and TAM of the subject and object to which they relate.
[9]For the ease of reading, we denote $\lambda_0, \lambda_1, ..., \lambda_5$ with $\lambda_{POS}, \lambda_G, ..., \lambda_{TAM}$, and $\lambda_6$ with $\lambda_L$.

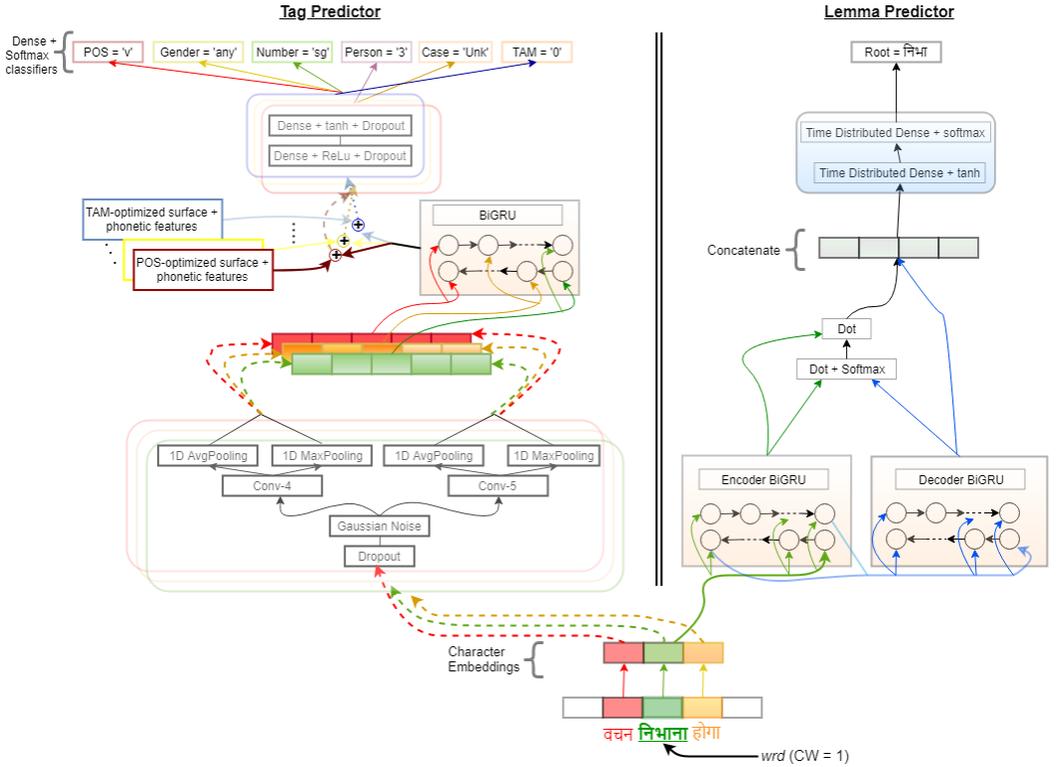

Fig. 1. Proposed MT-DMA framework, with an example for morphological analysis of the word निभाना [ni-BANA] (fulfill).

The tag predictor employs two regularization techniques (see Section 3.4) to prevent overfitting of the computed embeddings: (a) a dropout rate of 0.50 [78] and (b) a Gaussian noise layer with a mean of zero [63]. The noise-injected vectors are fed into separate sets of convolution and pooling layers (one set for each of the (2*CW)+1 words) that help capture short-range regularities (using filter widths of 4 and 5) among the characters of individual words. A bidirectional Gated Recurrent Unit (Bi-GRU) [20] layer combines together the convolutions of the current word with those in its context so as to process the dependencies over longer ranges across all such features. The representation obtained from the Bi-GRU is then branched off (with repetition) with each such branch merging separately into tag-specific hand-crafted features (for each of the six tags), selected from a pool of linguistic features via a multi-objective GA optimization. Each branch is processed by six separate stacks of two fully-connected/dense and dropout layers before being passed onto a final dense layer with softmax activation, which serves to classify from among the set of possible markers of the tag.

---
[10]We stick to a vocabulary size of 89 such characters thus encompassing all the alphabets - 52 for Hindi and 40 for Urdu, and digits of the language alongside common punctuation symbols.
[11]We fix the embedding dimension to 64 in order to avoid larger accuracy-time tradeoffs.

The lemma predictor leverages a Bi-GRU based encoder-decoder architecture [19], thus treating the problem as a sequence-to-sequence mapping between the characters of the input word and the lemma. Unlike the tag predictor feeding upon the context, the lemma predictor accepts only the current word as an input. We employ a global attention variant of the Luong model [50] to focus on the characters that are relevant to predicting the lemma of the current word. A beam search strategy with a width of four is then used to decode the best lemma out of all candidate solutions.

### 3.2 Tag predictor

**Convolution:** Inspired by the work of Wang et al. [77], we employ a CNN-RNN model for the sequential modeling problem of tag prediction. Following Zhang and Wallace [83], we employ two parallel convolutional layers of filter widths ($w_f$) 4 and 5 to extract n-grams character vectors of size $len_{max}$ - $w_f$ + 1, where $len_{max}$ is the length of the longest word in the training corpus, *i.e.*, 18 for both Hindi and Urdu; every word shorter than this are dummy padded. These layers act upon the character embeddings of a word to learn low-level translation invariant features. Let C[*, $i$ : $i + w_f$] be the $i$-th to the $i + w_f$-th column vectors of the character-level representation captured by the convolution operation. Then, the $i$-th element produced by the feature mapping $c$ of the convolutional layer $l$ with embedding dimension $d$ (in our case, d = 64), activation function $\sigma$ (ReLu), weight matrix $H \in \mathbb{R}^{d*w_f}$, and bias $b$ is given by:

$$c_l[i] = \sigma(\sum(C_l[*, i : i + w_f]H_l) + b_l) \qquad (2)$$

Simply put, each word is converted into a series of 4 and 5 ngrams (by applying convolution on the character embeddings for each word). Max pooling and average pooling operations are carried out in parallel upon the outputs of the convolution layers to create fix-sized vectors independent of the word length. The max pool operation helps the network to capture the most useful of local features captured by the convolution matrices while dropping the rest. Contrary to Collobert et al. [22], we observe that the consideration of context information in our task imparts influence on the morphological and semantic role of a given word and hence, introduce an average pooling operation in parallel to the max pooling to avoid the abrupt loss of such influences from the context. Each of the max and the average pool operations transform the feature map (i.e., outputs of the underlying convolutional layer) from size ($len_{max}$-$w_f$+1) to $[\frac{len_{max}-w_f+1}{2}]$. The output vector $p$ from either of these operations can be given by:

$$p = [p_1, p_2, ..., p_{\frac{len_{max}-w_f+1}{2}}] \qquad (3)$$

Further, we apply N kinds of matrix weights $\{H_1, H_2, ..., H_N\}$ to generate N such feature maps for max and average pool operations each:

$$P_{max} = P[p_{max}^1, p_{max}^2, ..., p_{max}^N] \qquad ; P \in \mathbb{R}^{[\frac{len_{max}-w_f+1}{2}]*N} \qquad (4)$$

$$P_{avg} = P[p_{avg}^1, p_{avg}^2, ..., p_{avg}^N] \qquad ; P \in \mathbb{R}^{[\frac{len_{max}-w_f+1}{2}]*N} \qquad (5)$$

The matrix pairs $[P^4_{max}, P^4_{avg}]$ and $[P^5_{max}, P^5_{avg}]$ of Equations (4) and (5), obtained from the max and average pool operations over each word *i* out of (2*CW)+1 words using convolutions of kernel sizes 4 and 5 are concatenated together to get a combined feature vector $Z_i$. Then, the vectors $Z_i$ obtained for *wrd* as well as for all its context words are arranged in sequence to obtain *word_contexts_seq$_{wrd}$*. More formally:

$$Z_i = \oplus([P^4_{max}, P^4_{avg}], [P^5_{max}, P^5_{avg}]) \qquad Z_i \in \mathbb{R}^{\frac{len_{max} - w_f + 1}{2} * (N*2)} \qquad (6)$$

$$word\_contexts\_seq_{wrd} = Z_1, Z_2, .., Z_i, .., Z_{(2*CW)+1} \qquad (7)$$

**Recurrent Neural Network:** *word_contexts_seq$_{wrd}$* is fed into an RNN as a sequence, to predict the tags for *wrd*. For the choice of RNN, we experiment with both Long Short Term Memory (LSTM) [33] and GRU units. Both the models perform equally well. However, the lack of controlled exposure of the memory content and independent control of the new candidate activation makes the training and inference of GRUs computationally cheaper[12] than LSTMs. On this account, we employ GRUs as our standard RNN variants. In particular, we employ a BiGRU to analyze the information flow from either directions.

*3.2.1 Linguistic features.* In order to better capture the language-specific information for Hindi, we incorporate a 54-dimensional feature set constituted by two classes: (a) *surface features* dealing with the context and morphology of the word, (b) *phonological features* dealing with the relationships among various speech phonemes. The complete list of these features can be found in Table 2.

Table 2. List of linguistic features

| Category | Sub-categories | Value descriptions |
|---|---|---|
| Surface | Length of the token (LoT) | - |
| | Position of the token | Boolean features specifying whether the token occupies (a) the first (is_first) or (b) the last (is_last) position in the sentence. |
| | Prefixes | The first one (pref-1), two (pref-2), and three (pref-3) characters of the token. |
| | Suffixes | The last one (suff-1), two (suff-2), three (suff-3), and four (suff-4) characters of the token. |
| | Context | The immediate successor (NW) and predecessor (PW) of the current token. |
| Phonological | Type-based | Total number of vowels, vowel-modifiers, consonants, punctuation marks, digits, halant, and nuktās (i.e., a diacritic in the form of a dot placed below a character). |
| | Aspirated consonant | Voiced (V) and voiceless (VL) aspirated consonants. |
| | Origin | Bangla (B), Dravidian (DV), and Devanagari (DN). |
| | Diphthongs (Is_diphthong) | - |
| | Place of Articulation (PoA) | Dental (D), Labiodental (LD), and Glottal (G). |
| | Modifier type | Anunaasika (AK), Anusvaara (AV), and Visarga (V). |
| | Height (for vowels) | Front (F), mid (M), and back (B). |
| | Length | Long (L), short (S), and medium (M). |
| | vowel (type-1) | Low (L), lower-middle (LM), upper-middle (UM), lower-high (LH), and high (H). |
| | vowel (type-2) | Samvrit (S), Ardh-samvrit (AS), Ardh-vivrit (AV), and Vivrit (V). |
| | Place | Dvayoshthya (DV), Dantoshthya (DN), Dantya (D), Varstya (V), Talavya (T), Murdhanya (M), Komal-talavya (KT), Jivhaa-muliya (JM), and Svaryantramukhi (SY). |
| | Manner | Sparsha (SP), Nasikya (N), Parshvika (PS), Prakampi (PK), Sangharshi (SN), and Ardh-svar (AS). |

---

[12]The same architecture when replaced with GRU units would take one-third the inference time of LSTM-based model (Appendix D).

For phonological features, we draw upon the work of Singh [68] that uses a number of modern, traditional Indian, and computation oriented phonological features that are particularly developed for languages employing the Brahmi script. These include a mixture of boolean and non-boolean features. The intuition behind incorporating phonological features comes from a number of works on the phono-morphological interface [10, 21] stating that the changes caused in a word's phonological well-formedness due to frequent morpheme assembly need to be addressed by the phonology.

Due to the unavailability of such thoroughly explored features for Urdu, we use the same phonological feature set assuming the relatedness of the Urdu etymology and grammar with that of Hindi, as described in Section 1.2. In fact, the International Phonological Alphabet (IPA) representation for Hindi and Urdu pronunciations is built upon the same underlying assumptions. We observed that the IPA's mappings[13] were detailed enough to identify the alphabet chart corresponding to the Urdu phonological features.

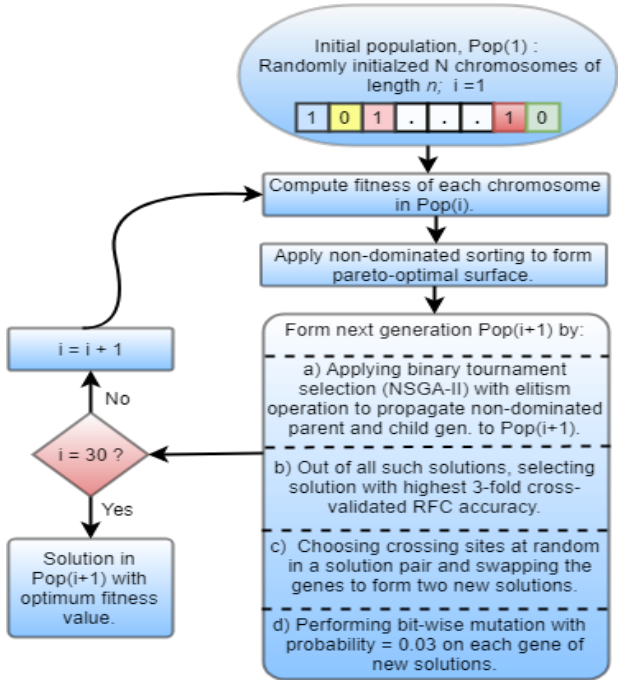

Fig. 2. Flowchart for the MOO technique: length *n* of each chromosome correlates with a subsample of the linguistic feature space, bits *1* and *0* denote the presence and absence of respective features.

**Multi-objective optimization for feature selection:** The complete set of features covered in Table 2 might not contribute equally in prediction for each tag. In order to avoid the additional complexities and loss of generalization arising due to the presence of additional features [71], we employ a multi-objective optimization (MOO) [64] using a genetic algorithm for the selection of linguistic features that are relevant to each of the tags. Following Akhtar et al. [2], the two objective functions used are: accuracy (maximize) and number of features (minimize). The parameters used were: number of generations (termination condition) = 30, population size = 60, crossover

---
[13]IPA Hindi-Urdu pronunciation mappings: https://en.wikipedia.org/wiki/Help:IPA/Hindi_and_Urdu

probability = 0.7, mutation probability = 0.03, and fitnesss function = test set accuracy (acc) of a Random Forest (RF) classifier[14]. Let, $U$ represent the set of all linguistic features and $M$ be the reward received on: (a) maximizing the accuracy of the RF classifier trained to predict tags and lemma using a subset of the features ($acc_{RF}(F)$), and (b) penalizing the number of phonetic features ($|F|$), in order to find the optimal subset $F^*$ from $U$. We simply choose $M$ to be the difference between the two. As encapsulated by equation 8, a separate $F^*$ is obtained for each tag by using a genetic algorithm to solve the optimization:

$$F^* = argmin_{F \in U}[|F| - acc_{RF}(F)] \qquad (8)$$

As depicted in Figure 2, we represent chromosomes using binary encoded strings. We train the Random Forest classifier model over the features represented by each chromosome, to predict the tags and the lemma, and use its micro-averaged F1 score obtained after 3-fold cross validation as the fitness value. Based on the evidence from Purshouse and Fleming [60], we use the elitism mechanism for quicker convergence of the population alongside maintaining consistent diversity and closeness among these by preserving the best string seen up to a generation. Upon termination, the best chromosome string contains the final set of tag-specific features to be used. To provide genetic diversity, we employ mutation operations on each member of the new solution. For a detailed explanation of MOO, readers are encouraged to refer to Kalyanmoy [39] and Hasanuzzaman et al. [32].

The tag-optimized linguistic features are concatenated together with the output vector of the Bi-GRU layer and fed to a stack of three dense layers - of sizes 64, 128 and $n_i$, with dropouts for each tag. The first dense layers employs ReLu activation, the second, tanh, and the last uses a softmax as the tag classifier.

### 3.3 Lemma predictor

The lemma predictor is essentially a seq2seq variant of the encoder-decoder model that shares the same embedding space as the tag predictor network to perform character-level transliteration of word. Unlike the tag predictor, the input to the lemma predictor model comprises only of the current word, and not its context. This design reflects our hypothesis that the gradual update of the embedding representations for the current word made by the tag predictor (which has access to the context) is eventually influenced by the context as the training progresses. As a result, the weight updates occurring within layers of the lemma predictor model exhibit influence of the context, despite having no direct access to the context. For reasons similar to those in Section 3.2, we choose single-layered BiGRUs as our encoding/decoding units. The encoder feeds upon the character embeddings of the current word. Because of the sequential processing of the characters by the encoder, the state of the encoder at one time step $t_h$ serves to capture a summary of all the character sequences that it has seen till now, i.e., $t_1, t_2, ..., t_{h-1}$. The decoder generates the lemma word by taking in as input the hidden states of the last time step of the encoder thus capturing the summary of the entire character sequence. We employ the Luong attention mechanism [50] with scaled energy term for allowing the decoder to attend to different parts of the source word at each step of the output generation. We also experiment with the popular Bahdanau [4] as well as the online monotonic alignments model [61], and obtain slightly overfitted results. The Luong model with the least parameter size was observed to generalize the best for our transliteration task.

---

[14]Parameter settings for the RF classifier: no. of trees = 15, split criteria = gini index, minimum no. of samples required to split each internal node = 2, minimum no. of samples required to be at a leaf node = 1, and no. of features for the best split = $\sqrt{n\_features}$, where n_features = no. of features in the gene pool for a generation.

### 3.4 Training apects

We jointly train both the components by minimizing the sum of the cross-entropy losses of all seven classifiers, weighted by their own contributions $\{\lambda_{POS}, \lambda_G, ..., \lambda_L\}$ as depicted in equation (1). Although, each classifier performed best with distinct contribution weights, we observed that the heuristic of alloting $\{\lambda_{POS}, \lambda_G, ..., \lambda_{TAM}\}$ the same value $\lambda$, while keeping $\lambda_L$ distinct, alleviated the overhead of precise adjustments. On obtaining an estimate for the value of $\lambda$ which gives optimal results for the majority of tag classifiers, we perform individual fine-tuning of the weight contributions for the sub-optimal ones by ensuring that the already optimal classifiers do not suffer any performance degradation.

For the tag predictor model, we found that its dependency upon the context words could make the model sensitive to the irregularities in the context (i.e., spaces, misplaced punctuation marks, and lack of enough context words in case of shorter sentences). In order to build resistance to such perturbations, we insert additive zero-centered Gaussian noise [63] into the embedded character inputs for all words of the tag predictor. The lemma predictor model (being context independent) did not require any such noise injections.

We use the Adadelta optimizer [82] with an initial learning rate of 1.0 for minimizing the joint loss function. We obtain significantly worse performance on using the standard Adam optimizer [43] equipped with a learning rate reducer factor of 0.2 after 5 epochs of patience monitored upon the validation loss. Further, we also experiment with using the Adam-to-SGD switching strategy, as mentioned by Keskar and Socher [41], and observe that the lemma predictions downgraded significantly, while the tags did not show much improvement upon that of Adam.

Finally, the joint training of the network eventually led to the early convergence of the tag predictors as compared to the lemma predictor. A model trained so, resulted in poor lemma qualities at the inference time despite good tag predictions. Thus, we adapt a progressive freezing strategy, as suggested by Brock et al. [11] to our training framework, by freezing the weights of all the layers of the tag predictor model along with the shared embedding layer, once their validation loss stops improving further. We then resume the training of the lemma predictor model by monitoring its validation loss. A more exhaustive freezing of only the tag predictor layers while allowing the shared embedding layer to contribute further to the backpropagation led to decrease in the accuracy of the tag predictions.

## 4 EXPERIMENTS

We evaluate MT-DMA on two separate datasets for Hindi and Urdu. This section describes the datasets, followed by a detailed discussion of the results and the errors encountered in predictions, from a linguistic perspective.

### 4.1 Datasets

We use the dependency treebanks of Hindi *(HDTB)* and Urdu *(UDTB)* [6] for our data. The Hindi database was downloaded from IIIT Hyderabad's website which at the time of writing, hosts the version 0.05 of the *HDTB pre-release*[15]. The HDTB database[16] is a collaborative research effort between five universities towards developing a multi-layered and multi-representational treebank for Hindi and Urdu. The treebank has sentences from conversations and news articles, each annotated with their true morphological tags and lemma, **in context**, as is shown in Table 3. We report our experiments on the 'news articles and cultural heritage' collections within the *CoNLL* section of the *Intra-chunk* corpus. Due to the unavailability of UDTB on the official website, we used the one hosted by the Universal Dependencies[17]. The fact that the Hindi and Urdu treebanks have each word manually annotated with the correct morphological tags and the correct lemma *in the contexts of the sentences the words occur in*, allows us to conduct us our experiments. These linguistic resources were highly useful for our task.

---
[15]http://ltrc.iiit.ac.in/treebank_H2014/
[16]http://verbs.colorado.edu/hindiurdu/

The Treebank is arranged by sentences, and each word of each sentence is annotated with the correct morphological analysis of that word in the context of the sentence. This is important for us, as this enables MT-DMA to be trained to be 'context aware', in that it takes into account not only a particular word but also its context words. As mentioned earlier, without the knowledge of context, multiple morphological analyses for a particular word can be arrived at, out of which only one is true in the context of the sentence.

| Tag | Sample classes | # possible classes ($n_i$) | |
|---|---|---|---|
| | | Hindi | Urdu |
| POS | Noun(n), Pronoun(pn), Adjective(adj), verb(v), adverb(adv), post-position(psp), avvya(avy) | 16 | 22 |
| Gender (G) | Masculine(m), Feminine(f), Any | 5 | 12 |
| Number (N) | Singular(sg), Plural(pl), Any | 4 | 5 |
| Person (P) | 1$^{st}$ (1), 2$^{nd}$ (2), 3$^{rd}$ (3), 3$^{rd}$ cardinal (3h), Any | 6 | 9 |
| Case (C) | Direct(d), Oblique(o), Any | 5 | 5 |
| TAM | है, का, ना, में, या, या1, कर, ता, था, को, गा, से, ए, 0 | 33 | 25 |

Table 3. Domain of tags

For all our experiments, we preserve the standard train-test-development splits of both the corpora which includes 80:10:10 ratio for train, test and development sets respectively. *The test set is held out separately, and is used only to report final results.* It is worth mentioning that the Urdu vocabulary size amounts to nearly one-third of Hindi, while having an average of five word longer sentences than the latter. We use an Unknown (Unk) marker to denote the missing annotations of a word for any tag. Table 4 provides a token-level statistics of both the data sets.

Table 4. Statistics of the HDTB and UDTB datasets

| Data | HDTB | | | | UDTB | | | |
|---|---|---|---|---|---|---|---|---|
| | Train | Dev. | Train + Dev. | Test | Train | Dev. | Train + Dev. | Test |
| # sentences | 14,089 | 1,743 | 15,832 | 1,804 | 5,599 | 612 | 6,211 | 620 |
| Mean length of sentences (# words) | 21.462 | 21.650 | 21.482 | 21.593 | 28.325 | 27.652 | 27.999 | 29.405 |
| # words | 302,375 | 37,736 | 340,111 | 38,954 | 158,589 | 16,923 | 175,512 | 18,231 |
| # unique words | 17,687 | 5,590 | 18,892 | 5,681 | 11,695 | 3,158 | 13,222 | 3,320 |
| # ambiguous words | 6,341 | 1,551 | 7,203 | 1,561 | 4,635 | 855 | 4,976 | 931 |

The following two statistical inferences can thus be built about the datasets from Table 3:

(1) While the HDTB dataset is far more larger than UDTB (1.9 times the total word count), Hindi possesses a greater proportion of ambiguous words - words with multiple valid analyses, when compared to Urdu.
(2) Sentences in Urdu are consistently longer (1.3 times) than those in Hindi.

---
[17]https://github.com/UniversalDependencies/UD_Urdu-UDTB/tree/master

### 4.2 Results

We evaluate MT-DMA's performance accuracy scores on each tag separately, as well as on different combinations of tags, and the lemma. For Hindi, we compare our results with two other functional analyzers hosted by: (A) IIT-Bombay (**IIT-B**)[18] and (B) IIIT-Hyderabad (**IIIT-H**)[19]. These are the state-of-art systems which are publicly available, and whose results can be reproduced on our test set. The test sentences were first fed manually into both the systems followed by downloading their respective outputs. For the purpose of evaluation, we consider unique candidate solution for each token in the test set, thus retaining the first candidate for tokens with multiple analyses. Due to the difference in symbology employed by IIT-B, its results further required mapping onto the symbol space leveraged by MT-DMA and IIIT-H prior to evaluation.[20] We use a baseline model (denoted by *"Baseline"*), composed of a bidirectional GRU layer followed by a dense layer with softmax activation, and train it independently on each of the seven tasks, thus aiding to compare the performance on individual taks. We employ two additional neural architectures (**char-CNN** and **Bi-GRU**) which are the adaptations of MT-DMA built upon the tag predictor: (a) char-CNN preserves all but the GRU layers of the tag predictor so that its framework is composed mainly of the CNN layers whose outputs are then concatenated with tag-specific phonetic features to be fed into the dense layers, and (b) Bi-GRU removes the CNN layers, so that the character level embeddings are fed directly to the Bi-GRU layers. Together, these adaptations serve to assist in comprehending the relative importance of CNN and GRU layers in the MT-DMA architecture while also providing additional neural baselines.

A comparison of IIT-B, IIIT-H, char-CNN, Bi-GRU and MT-DMA are provided in Table 5, wherein we evaluate MT-DMA's performance accuracy scores on each tag separately, as well as on different combinations of tags, and the lemma. The values reported in the table are simple 0-1 accuracies, which calculate the proportions of correct predictions out of the total predictions made. It is evident from the table that MT-DMA significantly outperforms existing analyzers for Hindi, as well as our baselines. Appendix C deals with detailed visualization of precision-recall scores for the individual tag classes while portraying the shared representations learned by the Bi-GRU layer of the Hindi tag predictor.

Additionally, while Table 5 calculates a binary accuracy of lemmas, we also compare the generated and target lemmas using Bilingual Evaluation Understudy Score (BLEU) [57] score, and Levenshtein distance, by treating both generated and target lemmas as sequences of characters. We obtain BLEU score of 93.240 (averaged over character 4-grams) and an average Levenshtein distance of 0.339 with respect to the actual lemmas.

In absence of such a functional analyzer for Urdu at the time of experiment, we set up the version 1.4 of the morph analyzer download available on the web page of Indian Language Technology Proliferation and Deployment Centre (TDIL).[21] Table 6 presents a detailed comparison between the accuracy scores of TDIL and MT-DMA. Further, for the predicted lemmas, we obtained a BLEU score of 86.836 and a Levenshtein distance of 0.376 against the corresponding actual lemma averaged over the test set.

It is worthwhile noting that two other systems appear in literature [51, 70]. However, neither is a) the code for these systems made publicly available, nor is b) the dataset they used available publicly. Hence, we make a direct comparison of our scores on the test sets[22] in lieu of reproducing their results (see Appendix E). These indicative results also depict that MT-DMA outperforms their results.

---

[18]http://www.cfilt.iitb.ac.in/~ankitb/ma/
[19]http://sampark.iiit.ac.in/hindimorph/web/restapi.php/indic/morphclient
[20]HDTB - the dataset used for training MT-DMA, and the IIIT-H system share common symbol space defined by the Language Technologies Research Centre, IIIT Hyderabad.
[21]http://tdil-dc.in/index.php?option=com_download&task=showresourceDetails&toolid=1522&lang=en
[22]They report their results on the Hindi [67] and Urdu Treebanks [7].

Table 5. Results for Hindi: best scores reported in bold across the experiments. MT-DMA is our best system.

| Analysis | Test data - overall (%) | | | | | |
|---|---|---|---|---|---|---|
| | Baseline | IIT-B | IIIT-H | char-CNN | Bi-GRU | MT-DMA |
| L | 72.55 | 72.30 | 77.48 | 75.17 | 79.02 | **82.29** |
| POS | 69.03 | 76.92 | 78.20 | 81.04 | 80.69 | **87.60** |
| G | 63.27 | 80.11 | 80.65 | 85.30 | 83.22 | **89.98** |
| N | 68.20 | 81.39 | 79.22 | 81.69 | 77.94 | **83.65** |
| P | 72.09 | 83.88 | 86.04 | 88.52 | 86.29 | **91.16** |
| C | 62.85 | 79.01 | 77.84 | 79.43 | 79.61 | **82.33** |
| TAM | 60.46 | 74.40 | 76.59 | 79.21 | 81.14 | **85.11** |
| $L + C$ | 60.99 | 68.13 | 73.70 | 72.09 | 75.95 | **78.02** |
| $G + N + P$ | 61.76 | 75.81 | 76.42 | 79.23 | 75.90 | **82.31** |
| $G + N + P + C$ | 58.44 | 73.87 | 72.96 | 77.06 | 73.25 | **85.30** |
| $L + G + N + P$ | 57.76 | 64.28 | 66.44 | 65.82 | 68.22 | **73.85** |
| $L + G + N + P + C$ | 55.61 | 62.10 | 64.58 | 62.97 | 65.84 | **70.49** |
| $L + POS + G + N + P + C + TAM$ | 50.35 | 52.60 | 55.76 | 59.91 | 57.66 | **64.31** |

Table 6. Results for Urdu: best scores reported in bold across the experiments. MT-DMA is our best system.

| Analysis | Test data - overall (%) | | | | |
|---|---|---|---|---|---|
| | Baseline | TDIL | char-CNN | Bi-GRU | MT-DMA |
| L | 68.01 | 64.07 | 68.52 | 72.18 | **76.91** |
| POS | 67.22 | 70.38 | 75.03 | 73.71 | **78.60** |
| G | 65.49 | 69.11 | 76.49 | 76.20 | **80.98** |
| N | 69.48 | 72.09 | 75.44 | 72.60 | **77.36** |
| P | 64.66 | 74.19 | 80.85 | 76.39 | **83.10** |
| C | 61.19 | 67.01 | 74.10 | 71.99 | **78.33** |
| TAM | 60.32 | 75.32 | 77.35 | 75.81 | **79.01** |
| $L + C$ | 59.17 | 62.19 | 66.71 | 69.59 | **75.22** |
| $G + N + P$ | 61.31 | 64.72 | 71.98 | 68.91 | **73.55** |
| $G + N + P + C$ | 57.71 | 61.08 | 68.22 | 66.85 | **69.98** |
| $L + G + N + P$ | 54.89 | 60.11 | 63.72 | 67.04 | **70.52** |
| $L + G + N + P + C$ | 52.11 | 58.44 | 60.54 | 64.88 | **67.19** |
| $L + POS + G + N + P + C + TAM$ | 48.05 | 48.26 | 52.45 | 55.13 | **61.31** |

### 4.3 Effect of Phonological and Lexical Features

Table 7 shows the accuracies for experiments performed under three conditions: a) using the complete set of phonological and surface features mentioned in Table 2, b) using the MOO-optimized subset of features for each tag (Section 3.2.1), and c) without the use of any features. Across every tag for both Hindi and Urdu, the evidence is in favor of using these features, and more specifically, in favor of using a (tag-dependent) optimized subset of these features rather than all of them. A complete set of the optimized features along with the population-wise accuracy of the Random Forest classifier can be found in Appendix B.

Table 7. F1[23] scores under various settings showing the effects of MOO-based feature optimization for Hindi and Urdu

| Tag | Hindi | | | Urdu | | |
|---|---|---|---|---|---|---|
| | Optimized | All | None | Optimized | All | None |
| POS | **99.30** | 96.60 | 93.17 | **96.30** | 93.71 | 89.64 |
| G   | **99.09** | 96.53 | 94.78 | **97.45** | 94.22 | 91.27 |
| N   | **99.24** | 95.81 | 92.75 | **97.83** | 93.78 | 91.96 |
| P   | **99.11** | 96.56 | 92.86 | **98.91** | 94.68 | 90.47 |
| C   | **99.03** | 95.97 | 93.45 | **97.04** | 94.96 | 91.06 |
| TAM | **99.98** | 96.84 | 94.22 | **99.19** | 96.77 | 92.50 |

### 4.4 Generalizability to Other Languages: Urdu

As is evident from Table 6, MT-DMA outperforms TDIL on Urdu. This is an attempt to study the extent to which our model is language agnostic. Further, Table 7 shows that the architecture as well as the phonological features remain compliant across languages, with the latter being restricted for languages that share the same phonological space. The morpho-phonology of Hindi and Urdu, largely derived from the long-spanning influence of Perso-Arabic vocabulary (which originated in the 11th century and evolved till the 19th century) on the development of their linguistic universals [59], borrows its words from several languages (Urdu from Persian and Arabic, and Hindi from Sanskrit). Despite such a distinct nature of borrowing, Hindi and Urdu appear to share a core vocabulary and grammar at the colloquial level owing to the cultural and historical similarities of the speakers [53]. However, at formal and literary levels, both the languages tend to diverge significantly in terms of lexicons (for e.g., basic words, tourist phrases, and school mathematics terms) [59], speech linguistics (loan word sounds, final consonant clusters, etc.) [55], orthographies (Hindi using the orthographically shallow alpha-syllabic Devanagari script while Urdu hosting the orthographically deep and phonologically opaque, Perso-Arabic script) [62], and word recognition methods (Hindi readers following the phonological assembly strategy while Urdu readers preferring the direct access route) [74]. In spite of such contrasting formal characteristics, the etymological and grammatical similarities of both the languages contribute to generalizing the linguistic features fine-tuned on one onto another.

### 4.5 Weight Calibration for Multi-task Learning

Similar to the technique followed by Dalvi et al. [26], we initially group the weights imparted upon all seven tags into two sets: $\lambda = \lambda_{POS} = \lambda_G = ,..., = \lambda_{TAM}$ for the tags, and $\lambda_L$ for the lemma. For smaller values of $\lambda$, all seven classifiers perform poorly. Further, assigning the value of $\lambda$ as zero reduces it to a single-task problem with the model optimizing the mere objective function for lemma prediction. The approach we use to arrive at the final values of $\lambda$s is to set $\lambda_L = 1-\lambda$, and vary $\lambda$ from 0 to 1, in increments of 0.1 to find the optimum value for $\lambda_L$ for which the lemma is predicted with the best BLEU score on the development set. This is depicted as the BLEU score line in Figure 3, for which the optimum value of $\lambda$ is 0.7. $\lambda_L$ is therefore, 0.3 (1 - 0.7). We then vary the sub-optimal values of $\lambda_{POS}, \lambda_G,..., \lambda_{TAM}$ in the neighborhood of $\lambda$, and find each of their optimum values (based on F1-scores) separately, as shown in Figure 3. The best BLEU scores for both the languages were achieved at $\lambda_L = 0.3$ which also produced the optimal F1 scores for all the tags but gender, person, and case. Following this, we explicitly adjusted the weights for the gender, person, and case predictors ($\lambda_G = \lambda_P = 0.9, \lambda_C = 0.95$) upto the threshold that does not impart any modifications over the optimal accuracies of rest of the tasks.

We derive two major conclusions from the plot in Figure 3: (a) the drop in the BLEU score of the predicted lemmas for smaller values of $\lambda$ suggests that the weights learned by the shared embedding layer while learning to predict tags contain useful information for lemma prediction as well, and (b) the common trend of increase in F1 scores of all six tags upon incrementing $\lambda$ depict that these are correlated and share mutual gains in the accuracy of one another. These conclusions also affirm our hypothesis that the inherent relatedness among these morphological tasks is exploited well by training them jointly, in a multi-task setting.

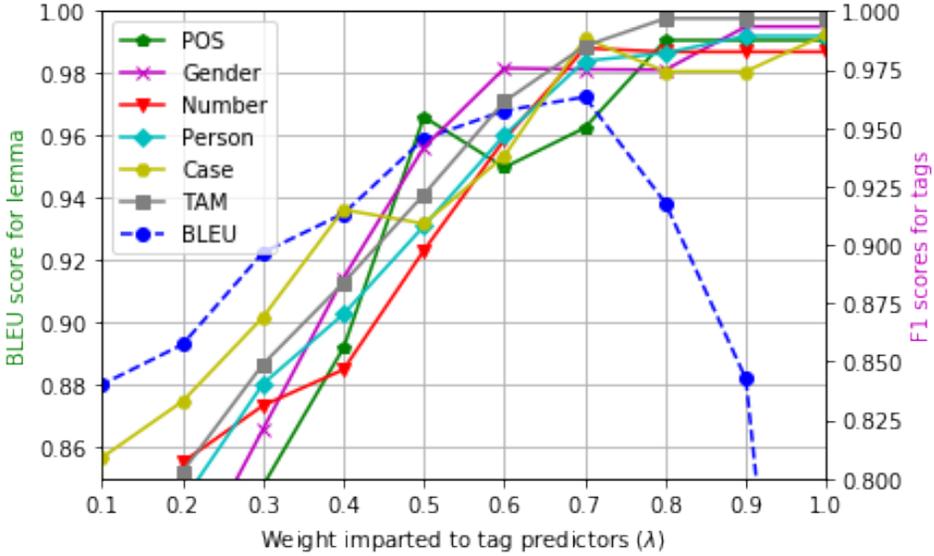

Fig. 3. lemma prediction vs. Morphological tagging weights for multi-task learning.

### 4.6 Comparison with Individual Learning of Tasks

We study the immediate effects of formulating the problem in a multi-task setting by training multiple individual models, each of which predicts an individual tag. This can be achieved by simply setting the weight contributions ($\lambda$s) of all but one tag as zeros. Figure 4 depicts the comparison between these settings for Hindi and Urdu. We obtain significantly better results by multi-tasking, with the average increase in F1 scores being 0.13 ( 15% increase) for Hindi and 0.17 ( 17% increase) for Urdu. These results, along with the conclusions from Section 4.5, strengthen the argument for a combined learning of the tags.

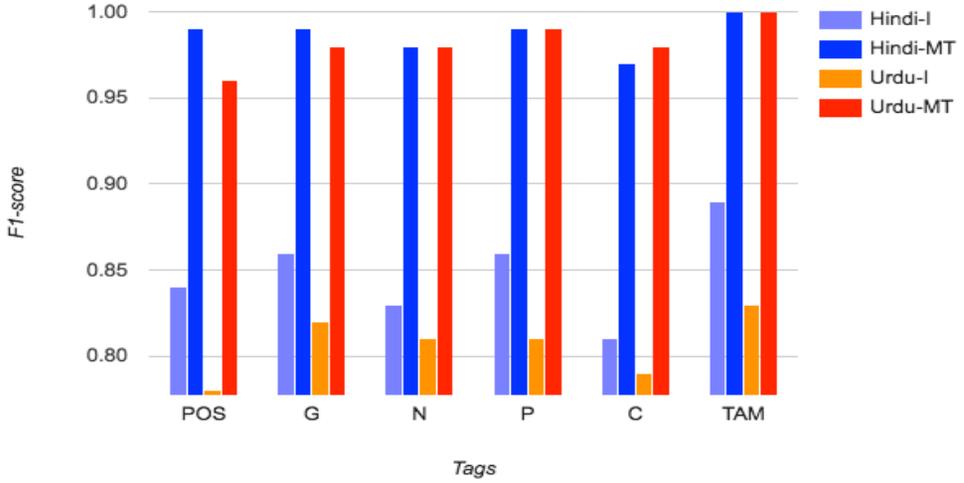

Fig. 4. Comparison of F1 scores for individual learning (I) i.e., without multi-tasking, and with multi-task learning (MT). For each tag, the four bars represent Hindi-I, Hindi-MT, Urdu-I and Urdu-MT in order.

### 4.7 Error Analysis

This section presents a summary of the errors observed in the predictions of the Hindi analyzer on the held-out test set.

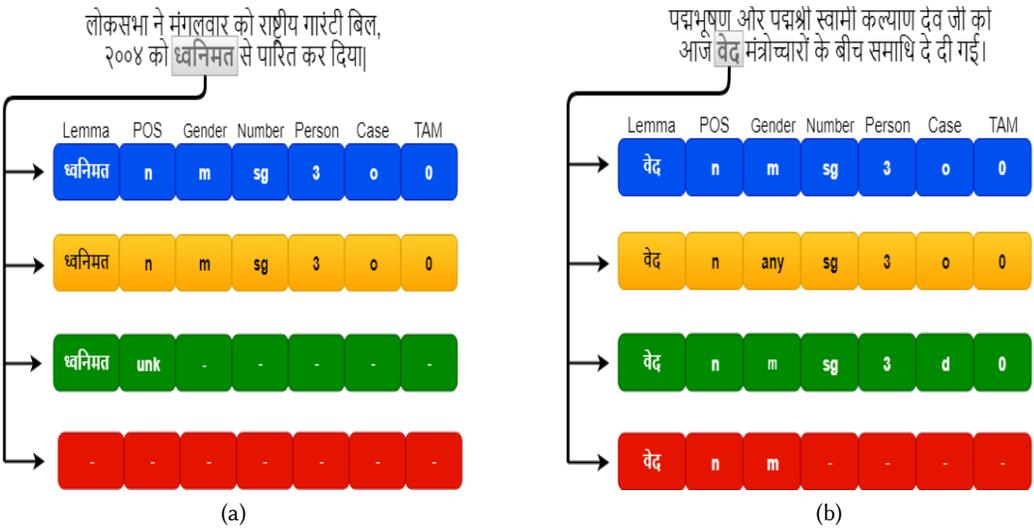

Fig. 5. Sample analyses of the three models on Hindi sentences showing mistaken and missing results; *color coding:* Ground truth values , MT-DMA , IIIT-H , and IIT-B .

Across all of the tags and lemma, the majority of errors arose when the word in question was borrowed from another language simply by transliteration. In our test set, this language was majorly English besides other Indic and Persian languages. The words transliterated from English in the test corpus accounted for an average of 45% of the total errors. इन्स्टिटचूट [instityUta] *(institute)* is a commonly occurring example of such a word. Further, another 9% of errors occurred in cases where the word itself was not an English word but the word in its immediate neighbour was. For example, in the phrase अनुसंधान इन्स्टिटचूट [anusanWAn instityUta] *(research institute)*, since इन्स्टिटचूट [instityUta] *(institute)* is an English word, the model does not have sufficient in-vocabulary context to analyse the Hindi word अनुसंधान [anusanWAn] *(research)*. As expected, the model performs poorly on named entities. Leaving POS out, mispredictions on the names of places, people, etc. account for around 20% of all the errors. Further, misspellings (which were deliberately left uncorrected to test the model's robustness) contribute towards 4% of all errors. Mispredictions on hyphenated words such as लहू-लुहान [lahU-luhAn] *(blood-drenched)* and अस्त-व्यस्त [aswvyasw] *(helter-skelter)* make up for 7% of the errors. On hyphenated words, the model often predicts the correct tags of one of the constituent subwords (that are separated by hyphens), instead of those of the hyphenated word as a whole. Below, we highlight more specific errors on the tasks.

*4.7.1 POS Errors.* Errors due to the model failing to use context for disambiguating words with more than one POS such as कार्यवाहक [kAryaVAhaka] *(noun: caretaker, adjective: acting/working as in 'acting president')* account for around 38% of POS errors. Even though numbers have a separate label, viz. *num* for POS tagging, when numbers are written as words instead of numerals, the model confuses these for nouns, viz. *n*. For instance, when '19th' is written in words as उन्नीस्वां [unnIsvAM] *(nineteenth)*, it gets tagged as a noun instead of a number. The same applies to numeric quantities, like तिहायी [wihAyI] *(thirds)*. As a whole, number-related errors consitute 21% of POS errors. Non-verbs ending with ना [nA] and being misclassified as verbs, account for another 16% of POS errors. For instance, रंजना [ranjanA] *(person's name)* and गुना [gunA] *(noun: fold, as in 'two-fold' and 'three-fold')* were declared to be verbs by the model. This is likely because most infinitive verbs in Hindi take the form *Lemma + ना [nA]*.

*4.7.2 Gender Errors.* On words whose gender can only be determined from context words (such as gender-universal nouns), the model performs poorly when the disambiguating evidence is outside the context window. For instance, in the sentence कृष्णा ने कहा कि वह आ जायेगी [krishNA ne kahA ki vaha A jAyegI] *(Krishna said she would come)*, 'Krishna' is not a gender-specific name, whose gender can be determined only from the very last word in the sentence जायेगी [jAyegI]. This is the major source of gender related errors and makes up over 65% of gender errors.

*4.7.3 Number Errors.* In Hindi, most masculine nouns that do not end with an आ [A] have the same plural and singular form. It is then entirely left to the context to disambiguate such cases. These contribute to one source of errors (27% of Number errors). Another source of errors are short auxiliary verbs such as है [hE], हैं [hEM], थे [We] and था [WA] *(is/are/were/was)* making up for 21% of errors. A third kind of Number error comes from the failure to distinguish between singular male deferential usage (for example, in the case of imperatives) vs. plural usage, unless the context specifically uses nouns indicating deference or number-related adjectives indicating plural for instance, in cases of pronouns like आप [Ap] *(you)*.

*4.7.4 Case.* One recognizable confusion of the model (occuring in 32% of Case errors) is between oblique singular and direct plural in cases when the lemma word is a masculine noun ending with an आ [A], since both these have the same inflected form, only distinguished by affixes attached to the lemmas of context words. These errors indicate scope for better interaction between the tag and the lemma predictors.

*4.7.5 Lemma (lemma) Errors.* Over 70% of the errors are due to character-level decoding issues: (a) recurrence of the last character of the actual lemma - घुस्पैठियााा [GusapETiyAAAA] instead of घुस्पैठिया [GusapETiyA] *(intruder)*, (b) an extra character appended at the end - बिना [binA] instead of बिन [bina] *(without)*, and (c) missing out the last character - दृष्टान् [xqRtAMw] instead of दृष्टांत [xqRtAn] *(exemplification)*. These errors provide scope for further being eliminated by carefully constructed post-processing edit rules.

## 5 RELATED WORKS

**Hindi:** In their work, Bharati et al. [5] present a paradigm based analyzer (PBA) as what can be traced as the first inflectional analyzer for Hindi, based upon a combination of paradigms and a lemma word dictionary. Following them, Goyal and Lehal [31] maintain a fully database-driven approach for analyzing inflectional morphology while Aswani and Gaizauskas [3] formulate a semi-autonomous rule-based approach to derive an inflected word's lemma form. Kanuparthi et al. [40] extend the work of Bharati et al. [5] to handle derivational forms by exploring derivational rules built upon suffixes, and using these to find majority properties of derived words of a suffix. On the other hand, Kumar et al. [48] use the Stuttgart FST tool to design inflectional and derivational rules for POS tagging. Addressing the low coverage of the aforementioned analyzers, Malladi and Mannem [51] introduce a Statistical Morphological Analyzer (SMA) trained on the Hindi treebank [67] that uses Moses [46] to predict the lemma, separate linear SVM classifiers (with a CW=1) to predict G, N, P, and C, and a set of heuristics utilizing POS tag sequences to predict the Vibhakti (V) and TAM of words. Building upon their work, the SMA++ system designed by Srirampur et al. [70] adapts a range of Indic language-specific machine learning features to obtain improved classification performances on L, G, N, P and C predictions for Hindi, Urdu, Tamil, and Telugu. Agarwal et al. [1] build an analyzer based upon lemma word and exception matching mechanisms to generate L, POS, G, N, and P.

**Urdu:** Despite the rigorous exploitation of Urdu morphology in the literature [37, 14, 13], relatively fewer works have been made towards building end-to-end Urdu morphological analyzers. Hussain [36] report an FST based approach to develop the first finite state analyzer for Urdu using the *Lexc* compiler for lexicon encodings. Humayoun et al. [35] design a Haskell-based functional morphology system that offers language dependent and independent parts. They use a lexicon building tool using a list of paradigms to extract 4,131 words for generating 496,976 word forms. As a part of the Parallel Grammar project [12] that aims to develop a unified grammar writing methodology for typologically different languages, Bögel et al. [8] build a finite state morphological analyzer based upon a cascade of finite state transducers. They use this to translate Urdu and Hindi scripts into a common ASCII transcription system by exploiting their structural similarity. Unlike for Hindi, FST based approaches to Urdu analyzers do not work well for a number of reasons such as the split-orthography tendency of the Persio-Arabic script in addition to infixation-based morphology computation.[24] Further, Srirampur et al. [70] exploit the lexical and grammatical similarities of Indian languages to adapt their proposed features for training SMA++ to predict L,G,N,P and C on Urdu, Tamil and Telugu treebanks [67].

**Multi-task assisted morphology learning:** In the recent years, good amount of work can be traced in leveraging morphology learning for improving the performance of a number of NLP tasks. Dalvi et al. [26] report improved MT results on German-English and Czech-English using a dual-task NMT system with morphological tag information injected into the decoder. The multi-task learning strategy performed superior to the joint-data learning technique that predicted word sequences and tags assuming no interdependence. Niehues and Cho [56] present a multi-task NMT system to produce coarse POS tags and named entities for the source language, viz. German as well as words in the target language, viz. English through the exploration of parameter sharing across the tasks. Following their work, Conforti et al. [23] explore a similar system to produce lemmas in the target language. They also use their tagger to produce rich target POS tags that are used as additional input in resource-constrained training examples, lacking source language text.

---
[24]http://www.ldcil.org/up/conferences/morph/presentations/shahid.pdf

Prior to this, García-Martínez et al. [29] had formulated a factored NMT system by extending the Bahdanau attention model [4] to predict the lemma of the target word and its corresponding factors (i.e., POS, G, N, P, and tense) at every step, followed by the combined post-processing of these to generate the target word. Cotterell and Heigold [24] presented a multi-task learning approach to cross-lingual morphological transfer by treating individual languages as separate tasks and training a joint model for all the tasks. The authors thus claim to be able to transfer morphology between related Indo-European languages by sharing the same vector space for character embeddings.

# 6 CONCLUSION

This paper introduces the MT-DMA, the first context-aware neural morphological analyzer for Hindi and Urdu. To this end, we exploit linguistic features optimized for each of the tags, using a multi-objective GA optimization technique. On both the languages, we achieve new state of the art. We plan to extend our work to non-Indic languages. We also look forward to training more morphological tasks within the same framework, and to generate a set of universal morphological embeddings which can then be transferred to other tasks too, just as the morphology-aware pretrained embeddings [9].

## A  TAG AGREEMENT PATTERNS FOR HINDI

For the specific case of Hindi, the modifier-head agreement pattern depicts the agreement of modifiers, including determiners and demonstratives to agree with their head noun in G, N, and C while the noun-verb agreement pattern shows finite verbs to be: (a) agreeing with the noun phrase if it is in the direct case, (b) neutral if there is no noun phrase in direct case. Further, as mentioned in Ambati et al. (2010), the case markers of syntactico-semantic relations in Paninian grammar

# B MOO-BASED FEATURE SELECTION

Table 8. Tag-wise optimized linguistic features obtained from MOO

| Tags | Hindi | Urdu |
|---|---|---|
| POS | ***Surface:*** LoT, is_first, pref-1/3, suff-1/2/3, PW/NW; ***Type:*** #punct, #digits; ***Aspirated:*** VL; ***Is_dipthong***; ***PoA:*** G; ***Height:*** M/B; ***Length:*** L/M; ***Type-1:*** LM/H; ***Type-2:*** AV/V; ***Place:*** DV/KT; ***Manner:*** SP/PK/SN; | ***Surface:*** LoT, is_first, is_last, pref-1, suff-2/3/4, NW; ***Type:*** #punct, #digits, #cons; Is_dipthong, ***PoA:*** G/LD; ***Height:*** M; ***Length:*** L/M; ***Type-1:*** H; ***Type-2:*** SN/AV; ***Place:*** DV/D; ***Manner:*** N/SP/PR/AS |
| Gender | ***Surface:*** is_last, suff-1/3, PW; ***Type:*** #vowels, #nuktas, #digits, #consonants; ***Aspirated:*** VL; ***Origin:*** DV; ***Height:*** M; ***Length:*** M/L; PoA: D/LD; ***Place:*** DV/V/M/KT; ***Manner:*** N/S/PK | ***Surface:*** is_first, is_last, suff-1/2/3, PW/NW; ***Type:*** #vowels, #digits, #nuktas; ***Aspirated:*** VL; ***Origin:*** DV; ***Height:*** M/B; ***Length:*** L; PoA: LD/D; ***Type-1:*** LM; ***Type-2:*** V; ***Place:*** DV/M/KT; ***Manner:*** N/PV/AS |
| Number | ***Surface:*** LoT, pref-1, suff-1/3, PW/NW; ***Type:*** #nuktas, #punctuations; ***Aspirated:*** VL; ***Origin:*** DV; ***Is_dipthong***; ***PoA:*** D; ***Height:*** F; Lenght: S/M; ***Type-1:*** LH; ***Type-2:*** S/AS/AV; ***Place:*** D/T; ***Manner:*** N/SP/PV/PK/AS | ***Surface:*** LoT, is_last, pref-1/2, suff-2, PW; ***Type:*** #vowels, #nuktas; ***Aspirated:*** VL; ***Origin:*** DV/B; ***Is_dipthong***; PoA: D/G; ***Height:*** M/B; ***Length:*** S; ***Type-1:*** LH/LM; ***Type-2:*** SN/AV/V; ***Place:*** D/T/KT; ***Manner:*** SP/PK/SN; |
| Person | ***Surface:*** is_first, pref-3, suff-3, PW/NW; ***Type:*** #vowels, #nuktas, #punct, #digits; ***Aspirated:*** VL; ***Origin:*** B/DN; ***Height:*** F/M/B; ***Length:*** S; ***PoA:*** D/G; ***Type-1:*** UM/LH/H; ***Type-2:*** S/AS/AV; ***Place:*** DV/D/V; ***Manner:*** SP/AS | ***Surface:*** is_first, is_last, suff-2/3/4, PW/NW; ***Type:*** #cons, #vowels, #punct #digits; ***Origin:*** B; ***Height:*** M/B; ***Length:*** L/S/M; ***PoA:*** G; ***Type-1:*** LM/UM/H; ***Type-2:*** AS/AV; ***Place:*** V/T/M; ***Manner:*** SP/PS; |
| Case | ***Surface:*** LoT, is_last, NW; ***Type:*** #nuktas, #punct, #digits; ***Aspirated:*** VL; ***Height:*** F/M; ***Length:*** L/S; Is_diphthong; ***Type-1:*** UM/LH; ***Place:*** DV/V/T/M/KT; ***Manner:*** N/SP/PV | ***Surface:*** is_first, PW/NW, pref-1; ***Type:*** #digits; ***Aspirated:*** VL; ***Height:*** F/B; ***Length:*** S/M; Is_diphthong; ***Type-1:*** L/LH/UM; ***Type-2:*** AS/V; ***Place:*** V/M/KT; ***Manner:*** SP/PS/PV |
| TAM | ***Surface:*** LoT, is_first, is_last, pref-1/2/3, suff-3/4, PW/NW; ***Type:*** #vowels, #nuktas, #digits; ***Aspirated:*** VL; Is_diphthong; ***Height:*** F; ***Length:*** L/M; ***Origin:*** DV/B/DN PoA: LD/D/G; ***Type-1:*** UM/LH/H; ***Type-2:*** AS/AV/V ; ***Place:*** D/T; ***Manner:*** SN | ***Surface:*** LoT, is_last, pref-3, suff-2/3/4, NW; ***Type:*** #vowels, #nuktas; ***Aspirated:*** VL; Is_dipthong; ***Height:*** F/M; ***Length:*** L/M/S; ***Origin:*** DV/B; ***PoA:*** LD/D/G; ***Type-1:*** LH/H; ***Type-2:*** SN/AS; ***Place:*** T/KT; ***Manner:*** PS/PK |

## C.1 Precision-Recall curves for individual tag classes

## C RESULT VISUALIZATIONS

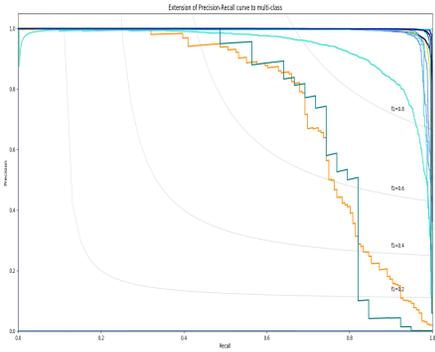

(a) POS

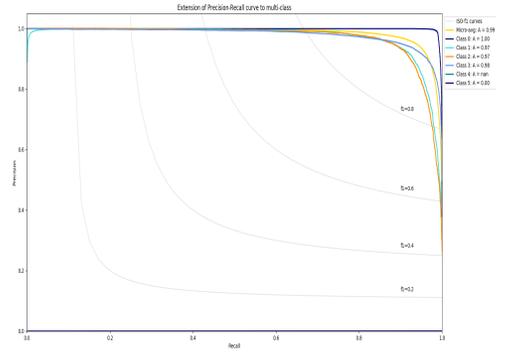

(b) Gender

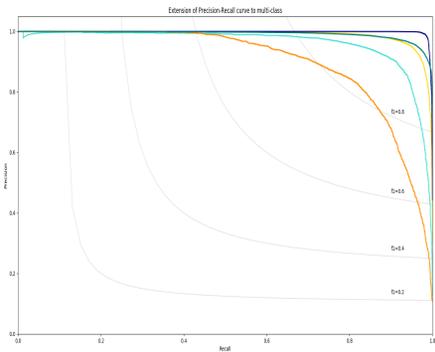

(c) Number

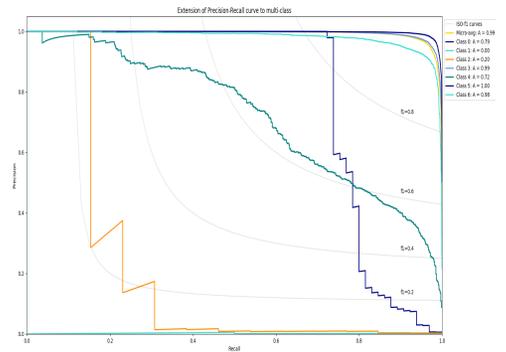

(d) Person

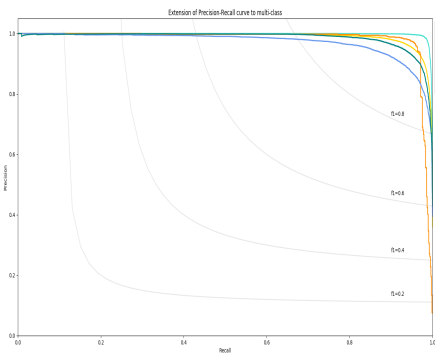

(e) Case

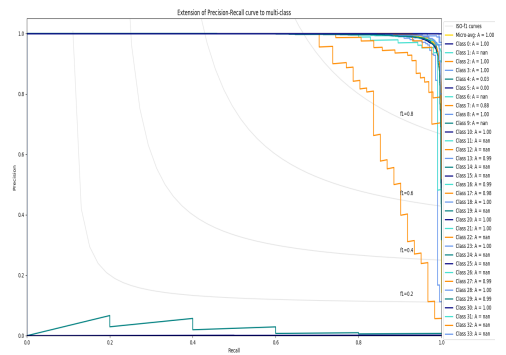

(f) TAM

Fig. 6. Micro-averaged Precision curves arranged by increasing F1 scores for individual classes of tags.

## C.2 t-SNE visualization of Tag representations

*Interpretation:* Each colored point in the two-dimensional space corresponds to an individual marker within the tag class, *i.e.*, the distances between classes in the input space is preserved in the two dimensional space.

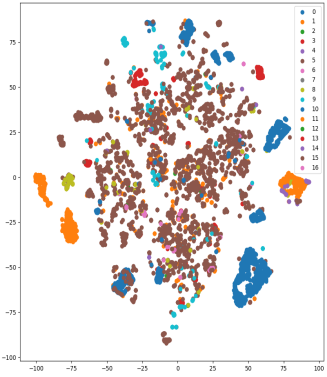

(a) POS: D=73.129, E=0.769

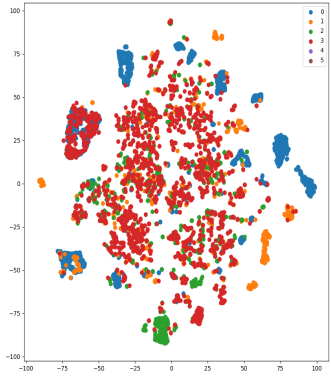

(b) G: D=72.99, E=0.771

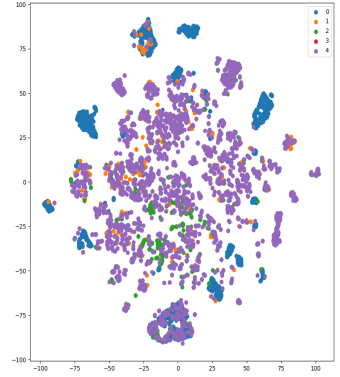

(c) N: D=72.868, E=0.769

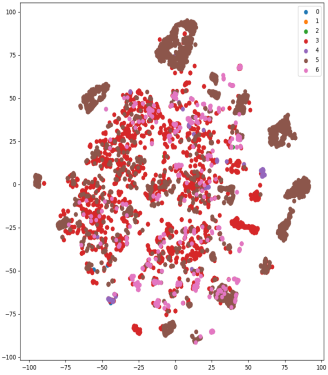

(d) P: D=73.069, E=0.763

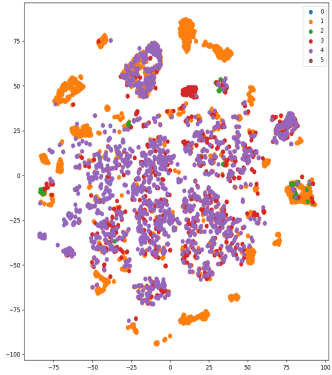

(e) C: D=73.119, E=0.768

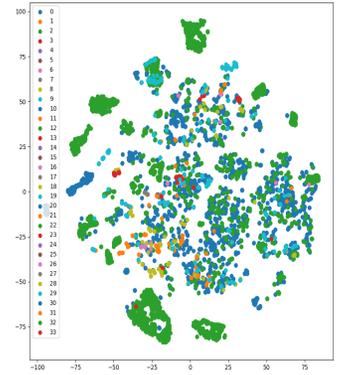

(f) TAM: D=72.919, E=0.774

Fig. 7. t-SNE visualizations of the representations learned by the Bi-GRU layer common to all six tag predictors: *D* denotes the K.L. divergence values and *E*, the errors over 10,000 random instances of the HDTB test dataset.

## C.3 Test set accuracy of Random Forest (RF) Classifiers

*Interpretation:* The non-linear sigmoid growth curves across all tag markers depict the increase in generalization capability [25] of the individual instances of RF classifiers as the proportion of the included population grows. The classifier can be seen to deliver best results on TAM marker while the Case (C) marker offers least generalization.

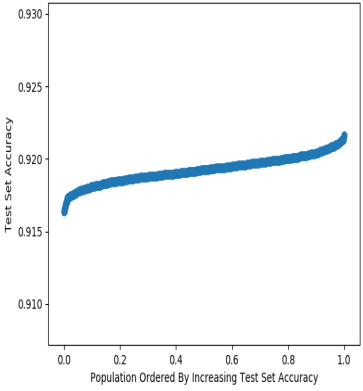

(a) POS: $\alpha$ = 91.94%

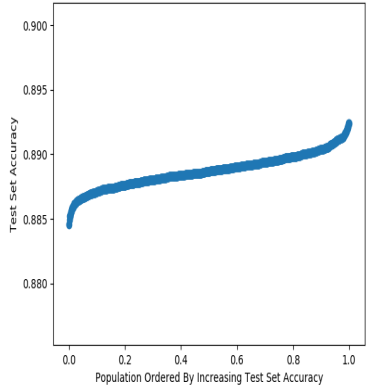

(b) G: $\alpha$ = 88.87%

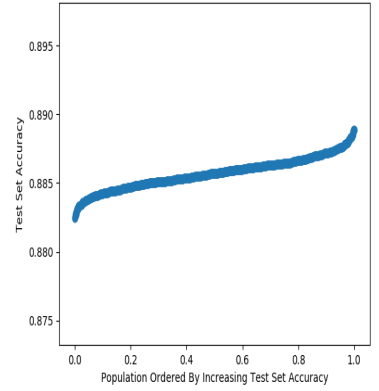

(c) N: $\alpha$ = 88.62%

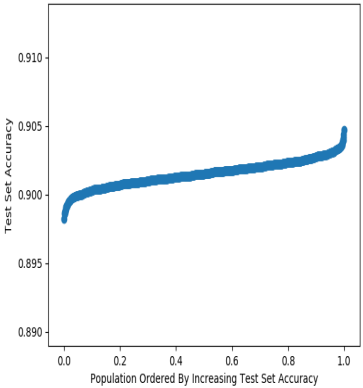

(d) P: $\alpha$ = 90.17%

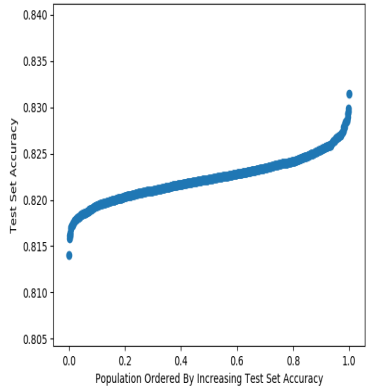

(e) C: $\alpha$ = 82.27%

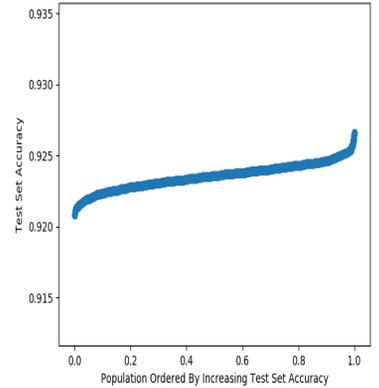

(f) TAM: $\alpha$ = 92.39%

Fig. 8. Test set accuracy vs. Percentile of population ordered by increasing accuracy scores on the HDTB test dataset: $\alpha$ denotes the test set accuracy averaged over all 30 generations of the population.

# D EXPERIMENT BUILD-UP

Table 9. Results for experimental settings with various architectures: *the expriments were performed in the top-down order; the ditto (" ") marks denote the architecture inheriting similar settings form the row above in addition to the new changes listed.*

| Configurations | POS | G | N | P | C | TAM | Epochs for Convergence | Inference time (seconds) |
|---|---|---|---|---|---|---|---|---|
| A. With simple Bi-LSTMs | | | | | | | | |
| BLSTM, CW = 1 | 0.74 | 0.70 | 0.77 | 0.81 | 0.81 | 0.66 | 21 | 105 |
| " " + attention | 0.77 | 0.71 | 0.78 | 0.83 | 0.83 | 0.79 | 26 | 108 |
| BLSTM, CW = 2 | 0.76 | 0.71 | 0.76 | 0.82 | 0.85 | 0.81 | 12 | 147 |
| " " + attention | 0.76 | 0.70 | 0.78 | 0.81 | 0.83 | 0.72 | 15 | 145 |
| BLSTM, CW = 3 | 0.77 | 0.71 | 0.78 | 0.81 | 0.85 | 0.81 | 10 | 182 |
| " " + attention | 0.78 | 0.75 | 0.77 | 0.83 | 0.83 | 0.79 | 14 | 177 |
| BLSTM, CW = 4 | 0.79 | 0.75 | 0.80 | 0.82 | 0.85 | 0.68 | 10 | 210 |
| " " + attention | 0.76 | 0.74 | 0.77 | 0.80 | 0.84 | 0.78 | 10 | 215 |
| BLSTM, CW = 5 | 0.74 | 0.76 | 0.78 | 0.82 | 0.76 | 0.71 | 11 | 243 |
| " " + attention | 0.74 | 0.77 | 0.79 | 0.82 | 0.76 | 0.74 | 10 | 255 |
| B. With CNN-RNN (RNN = GRUs) | | | | | | | | |
| CNN-RNN with maxPool (M), CW = 4 | 0.75 | 0.77 | 0.83 | 0.84 | 0.82 | 0.82 | 8 | 70 |
| CNN-RNN with avgPool (A), CW = 4 | 0.75 | 0.77 | 0.81 | 0.82 | 0.86 | 0.83 | 8 | 70 |
| " " + Gaussian noise | 0.87 | 0.87 | 0.92 | 0.93 | 0.93 | 0.93 | 15 | 71 |
| " " + AM pool, CW = 4 | 0.90 | 0.88 | 0.91 | 0.93 | 0.91 | 0.91 | 22 | 70 |
| C. With linguistic features | | | | | | | | |
| " " + all features | 0.91 | 0.89 | 0.93 | 0.94 | 0.92 | 0.94 | 23 | 69 |
| " " + GA optimized features | 0.97 | 0.94 | 0.96 | 0.96 | 0.96 | 0.97 | 16 | 52 |
| D. With attention-based lemma predictor | | | | | | | | |
| " " AM pool + Bahdanau attention | 0.96 | 0.92 | 0.95 | 0.96 | 0.95 | 0.98 | 16 | 59 |
| " " AM pool + Luong attention | 0.98 | 0.93 | 0.96 | 0.99 | 0.98 | 0.99 | 24 | 41 |
| " " + progressive freezing | 0.99 | 0.99 | 0.99 | 0.99 | 0.99 | 1.00 | 143 | 46 |

# E  DIRECT COMPARISON WITH PREVIOUS LITERATURE

Table 10. Comparison for Hindi

| Analysis | Test data - overall (%) | | | | Test data - OOV words (%) | | | |
|---|---|---|---|---|---|---|---|---|
| | O-PBA | SMA | SMA++ | MT-DMA | O-PBA | SMA | SMA++ | MT-DMA |
| L | 86.69 | 95.70 | 98.43 | **99.27** | 82.48 | 85.82 | 93.07 | **96.22** |
| POS | – | – | – | **99.06** | – | – | – | **89.61** |
| G | 79.59 | 95.43 | 96.21 | **99.33** | 44.06 | 79.09 | 83.11 | **91.85** |
| N | 80.50 | 94.90 | 95.47 | **98.25** | 47.56 | 89.12 | 92.81 | **95.04** |
| P | 84.13 | 95.77 | 96.28 | **98.93** | 53.89 | 94.39 | **96.17** | 94.92 |
| C | 81.20 | 94.62 | 95.43 | **97.41** | 47.36 | 87.40 | 89.45 | **91.39** |
| TAM | 59.65 | 97.04 | – | **99.68** | 34.56 | 96.04 | – | **88.64** |
| $L+C$ | 72.06 | 90.67 | 94.01 | **96.25** | 44.66 | 75.33 | 82.92 | **86.04** |
| $G+N+P$ | 73.81 | 89.42 | 90.36 | **94.70** | 39.58 | 71.31 | 77.24 | **83.19** |
| $G+N+P+C$ | 70.87 | 85.56 | 88.51 | **91.53** | 35.95 | 64.64 | 72.36 | **86.23** |
| $L+G+N+P$ | 66.28 | 85.88 | 89.26 | **91.41** | 38.46 | 62.34 | 72.82 | **79.51** |
| $L+G+N+P+C$ | 63.41 | 82.16 | 85.87 | **88.12** | 34.89 | 56.66 | 65.96 | **71.37** |
| $L+POS+G+N+P+C+TAM$ | 42.80 | – | – | **79.55** | 14.51 | – | – | **65.02** |

Table 11. Comparison for Urdu

| Analysis | Test data - overall (%) | | | Test data - OOV words (%) | | |
|---|---|---|---|---|---|---|
| | M | SMA++ | MT-DMA | M | SMA++ | MT-DMA |
| L | 93.65 | 95.34 | **97.49** | 87.54 | 89.21 | **92.80** |
| POS | – | – | **96.49** | – | – | **88.35** |
| G | 90.39 | 93.79 | **97.80** | 79.40 | 90.35 | **93.74** |
| N | 92.38 | 95.66 | **98.13** | 85.36 | **94.50** | 93.31 |
| P | 93.93 | 97.07 | **99.45** | 86.56 | **98.39** | 95.16 |
| C | 87.99 | 90.92 | **97.72** | 76.08 | 84.07 | **91.66** |
| TAM | – | – | **99.61** | – | – | **93.01** |
| $L+C$ | 82.94 | 86.93 | **91.17** | 67.25 | 75.66 | **87.04** |
| $G+N+P$ | 84.52 | 89.43 | **93.96** | 70.32 | 86.09 | **88.20** |
| $G+N+P+C$ | 77.01 | 82.17 | **87.71** | 58.54 | 73.69 | **83.58** |
| $L+G+N+P$ | 80.12 | 86.07 | **87.16** | 64.14 | 78.93 | **86.26** |
| $L+G+N+P+C$ | 73.11 | 79.16 | **82.88** | 53.30 | 67.98 | **74.19** |
| $L+POS+G+N+P+C+TAM$ | – | – | **74.65** | – | – | 68.90 |